\documentclass[11pt,a4paper]{article}

\usepackage[utf8]{inputenc}
\usepackage[T1]{fontenc}
\usepackage{amsmath,amssymb,amsthm}
\usepackage{graphicx}
\usepackage[colorlinks=true,linkcolor=blue,citecolor=blue,urlcolor=blue]{hyperref}
\usepackage[numbers]{natbib}
\usepackage{geometry}
\usepackage{booktabs}
\usepackage{caption}

\newcommand{\AneqAapproxA}{A \neq A \approx A}

\title{\textbf{NRR-Core: Non-Resolution Reasoning as a Computational Framework for Contextual Identity and Ambiguity Preservation}}

\author{
  Kei Saito\thanks{ORCID: 0009-0006-4715-9176} \\
  Independent Researcher, Japan \\
  \texttt{kei.saito.research@gmail.com}
}

\date{First posted 15 December 2025; revised July 2026 \\[0.5em]
\textit{Part of the Non-Resolution Reasoning (NRR) research program.}}

\begin{document}

\maketitle

\begin{center}
\textcopyright\ 2025 Kei Saito.
Licensed under CC BY 4.0.\\
\url{https://creativecommons.org/licenses/by/4.0/}
\end{center}

\begin{abstract}
Ambiguity loss is a persistent concern in language-processing systems that optimize for a single resolved output.
When context is incomplete, competing interpretations can be compressed too early into one response state.
Building on prior ambiguity-preserving and information-state approaches, we specify
\textbf{Non-Resolution Reasoning (NRR)} as an explicit retention--commitment
interface. NRR organizes context-indexed alternatives, independently active weights,
declared retention and commitment operations, and non-destructive output projection
around three principles: Context-indexed Non-Identity, Approximate Identity, and
Non-Resolution. The paper specifies a retained state and candidate operator vocabulary,
and proposes Multi-Vector Embeddings, Non-Collapsing Attention, and Contextual Identity
Tracking as implementable architectural realizations.
In a reproducible synthetic two-turn task, one gated Multi-Vector-Embedding instantiation maintains high output entropy before disambiguating context arrives ($H = 0.91$ bits, near the $1.0$-bit maximum), while a controlled single-embedding baseline has low entropy ($H = 0.15$ bits); both tested systems resolve correctly after context arrives.
This establishes that high pre-context output uncertainty and accurate later resolution
can coexist in the tested gated configuration. It does not validate the full NRR
architecture or matched-parameter superiority. The specification, proposed components,
and demonstrated behavior are distinct contribution layers. NRR targets premature
commitment, not commitment itself: alternatives can remain available while evidence is
incomplete, and commitment occurs at explicit output or action gates.
\textbf{The question is not whether AI should resolve ambiguity, but when, how, and under whose control.}

\textbf{Implementation}: A reference implementation accompanying this work 
is available at \url{https://github.com/kei-saito-research/nrr-core}.\\
\textbf{Series hub}: A cross-paper index for the NRR series is available at
\url{https://github.com/kei-saito-research/nrr-series-hub}.
\end{abstract}


\section{Introduction}

This paper develops the foundational layer of the Non-Resolution Reasoning (NRR) framework. It focuses on the criterion that ambiguity should remain available when context is still incomplete, and it should be read alongside companion papers that address text-to-state mapping, implementation, operational calibration, and bounded verification.

\subsection{The Problem of Premature Semantic Collapse}

Large language models (LLMs) have achieved remarkable success in natural language processing \cite{devlin2019bert, radford2018improving}, demonstrating fluency in text generation, translation, and question-answering.
At the same time, many deployed language systems are optimized to produce one resolved output at each step, which creates pressure toward early resolution when context is incomplete.
This ``premature semantic collapse bias'' manifests as:

\begin{itemize}
\item Binary decisions forced too quickly
\item Collapse of multiple valid interpretations into single outputs
\item Overconfident inference in uncertain contexts
\item Repeated full branchwise comparative evaluation during retention, followed by backtracking after premature fixation
\item Brittleness when handling paradoxes and self-referential structures
\end{itemize}

Autoregressive systems must realize an output, but output selection does not by itself establish that alternative interpretations have been erased from the internal state. We therefore use \emph{collapse} operationally for the tested state or behavior, not as a universal mechanistic diagnosis. Standard embedding layers and contextual encoders provide different forms of representation \cite{peters2018deep}; softmax attention normalizes weights competitively \cite{vaswani2017attention}; generation objectives optimize realized sequences; and decoding strategies can introduce output diversity \cite{holtzman2020curious, li2023contrastive}. None of these properties alone proves semantic erasure.

These observations motivate ambiguity retention as a design target rather than a default error condition.
They suggest that some failures around ambiguity, paradox, creativity, and nuanced reasoning may stem not only from data or scale limits, but also from \textbf{architectural assumptions about how meaning should be processed}.

Efficient attention, mixture-of-experts, probabilistic programming, and neurosymbolic systems provide other forms of parallelism or uncertainty representation \cite{child2019generating,dao2022flashattention,shazeer2017outrageously,goodman2008church,garcez2020neural}. NRR's specific design objective is to make the retention of multiple context-indexed interpretations and the transition to commitment explicit and inspectable; comparative advantages over those alternatives remain to be tested.

\subsection{Relation to Ambiguity-Aware and Interactive NLP}

The motivation for preserving ambiguity is not new to NRR. Packed representations and underspecified semantics have long represented several analyses without enumerating or resolving all of them immediately \cite{emele1998packed,alshawi2011underspecified,manshadi2018coherence}. Information-state dialogue management explicitly couples an information state with update rules and dialogue control \cite{larsson2000information}, while POMDP dialogue systems maintain uncertain dialogue-state hypotheses and select actions through a policy \cite{williams2007pomdp}. AmbigQA requires every plausible answer and a disambiguated question for each answer \cite{min2020ambigqa}. More recent work showed that language models often answer ambiguous questions instead of seeking clarification \cite{kuhn2023clam}, introduced benchmarks for recognizing and disentangling multiple sentence meanings \cite{liu2023ambient}, found weak uncertainty when models interpret semantically underspecified sentences \cite{wildenburg2024dust}, aligned models to detect and manage perceived ambiguity \cite{kim2024apa}, generated missing interpretations before semantic parsing \cite{saparina2025disambiguate}, and developed uncertainty- or intent-sensitive policies for asking clarification questions \cite{testoni2024asking,zhang2025clarify}.

These state/update/control traditions are direct predecessors to any generic claim that explicit state, update, or state-to-action control is new. NRR's narrower proposed synthesis uses context-indexed retained records, independently active non-normalized weights, a declared retention--commitment operator vocabulary, and a projection contract that does not itself delete the retained state.

Related work also studies how conclusions change over time. Belief-R evaluates whether models revise earlier inferences when later premises warrant an update, while also testing whether they avoid unnecessary revision \cite{wilie2024belief}. An arXiv preprint first posted in May 2025, and later published as an ICLR 2026 Outstanding Paper, finds that models often make early assumptions, attempt final solutions prematurely, and then fail to recover \cite{laban2026lost}. In referentially ambiguous dialogue, models tend to commit to one interpretation or enumerate all candidates instead of hedging or asking for clarification, and a simplified-language instruction can further reduce clarification behavior \cite{ellinger2025depends}.

Work appearing after the first NRR-Core preprint has converged even more directly on this problem family. It has named the epistemic risk of singularly resolving genuinely multiply interpretable terms as \emph{ambiguity collapse} \cite{gurarieh2026collapse}, measured premature closure in clinical response decisions \cite{handler2026closure}, tracked early diagnostic commitment and later self-correction \cite{fang2026mint}, studied premature confidence within reasoning trajectories \cite{gai2026confidence}, and proposed hidden-state diagnostics of representational commitment in agents \cite{mehta2026commitment}. These studies differ in construct, access regime, and task, but they make clear that premature commitment is a growing research area rather than a phenomenon uniquely identified by NRR.

This convergence makes the retention-to-commitment transition a timely engineering target. Core addresses that target at the architectural interface: it specifies what a system retains, where commitment occurs, and how a task-facing output can be separated from the retained state that remains available for later revision.

Accordingly, this paper does not claim the first recognition or preservation of linguistic ambiguity, explicit information or belief state, state update, state-to-action control, clarification, belief revision, or premature commitment. Its contribution is the integration and explicit specification of context-indexed identity, independently active interpretation weights, retention and commitment operators, and non-destructive output projection around a common retention--commitment interface. The integration is the proposed contribution; novelty of every constituent mechanism and comparative superiority over predecessor architectures are outside the present claim.

\subsection{Contributions of This Work}

This paper proposes Non-Resolution Reasoning (NRR), a computational framework that treats non-resolution---the deliberate maintenance of undecided states---as a valid reasoning mode.
Our contribution layers are:

\begin{enumerate}
\item \textbf{Foundational specification}: We organize context-indexed ambiguity retention around three principles, define a retained state with independently active weights, introduce a candidate state-level operator vocabulary, and separate task-facing projection from destruction of retained state.
\item \textbf{Proposed architectural realizations}: We propose Multi-Vector Embeddings, Non-Collapsing Attention, and Contextual Identity Tracking as implementable mechanisms, and provide benchmarking and cost considerations for their extension.
\item \textbf{Demonstrated implementation point}: Under one controlled two-turn task, one gated NRR-lite configuration maintains Turn 1 output entropy of $H = 0.91$ bits while the controlled single-embedding baseline has $H = 0.15$ bits across five random seeds; both tested systems resolve correctly after context arrives.
\end{enumerate}

The positive center of the paper is an explicit retention--commitment interface and a reproducible demonstration that one gated instantiation maintains high pre-context output entropy while resolving accurately after context arrives.

NRR is anti-premature-collapse, not anti-collapse. Commitment remains necessary, but its timing should be condition-designed rather than fixed by default: commitment should be an explicit, gated operation rather than an untracked consequence of representation and decoding. This concerns decision structure rather than interpersonal tone; outputs may remain concise/direct while preserving conditional optionality until fixation is warranted. NRR is a derivation-and-evaluation framework, not a closed recipe, and the same derivation lens supports additional operator/policy instantiations under fixed protocols with condition-bounded reporting.

Unsupported interpretations are not injected at a fixed evidence state. Hypothesis-set expansion is evidence-gated, unobserved possibilities remain unresolved, and evidence in this paper is restricted to explicit input, provided context, and declared retrieval outputs. In the series' narrow non-evaluative read, \textbf{meaning does not need to collapse to be computationally useful}: retained state should not default to repeated full branchwise comparative evaluation during retention, and commitment is deferred to explicit output or action gates.

\paragraph{Scope and series alignment.}
In this paper, $\sigma$ and $\delta$ denote the \emph{state-level} operator family in the foundational formalism; later implementation papers use item-level update operators ($\sigma_{\text{item}}, \delta_{\text{item}}$). Core evaluates foundational non-collapse metrics, especially Turn-1 entropy under ambiguity, while later papers add interface and operational metrics such as extraction reliability, token efficiency, and stability. These metrics are complementary layers rather than competing definitions of success. The empirical claim in this paper is limited to the controlled synthetic experiment and the formalization stated here.

\section{The NRR Model: Foundational Principles}

\subsection{The $\AneqAapproxA$ Principle}

NRR adopts a context-indexing convention: distinct occurrences of the same symbol may receive distinct representations while retaining a task-defined similarity relation. This does not reject the logical law of identity; it distinguishes symbol type from context-indexed occurrence and representation. We summarize this representational convention as $A \neq A$, yet $A \approx A$.
\textbf{Positional Non-Identity.}
If two tokens occupy different positions or contexts, they are treated as non-identical
($A_i \neq A_j$), even when they share the same symbolic form ($A_i \approx A_j$).
This divergence is not noise but signal: preserving positional difference is what makes structural similarity computable.
In NRR, $\neq$ marks contextual identity separation, while $\approx$ preserves structural/symbolic similarity.
The subscript notation ($A_i, A_j$) serves as shorthand for the functional form $I(A, c_i), I(A, c_j)$.
\textbf{Modeling contrast used here:} a context-invariant token-type convention assigns one representation to a symbol type, whereas NRR permits distinct representations for its context-indexed occurrences. This is a representational design contrast, not a contrast with the logical law of identity.
\textbf{NRR Formalization:}
We formalize this as a dual-condition relation:
\begin{equation}
I(A, c_1) \neq I(A, c_2) \quad \text{and} \quad I(A, c_1) \approx I(A, c_2)
\end{equation}

where $I(A, c)$ denotes the contextual identity of symbol $A$ in context $c$, and $\approx$ denotes operational equivalence under a task-specific similarity metric $S(\cdot,\cdot)$ with threshold $\tau$ (i.e., $I(A, c_1) \approx I(A, c_2)$ iff $S(I(A, c_1), I(A, c_2)) \ge \tau$).
In this framework, task-output validity is assessed under declared context or conditions rather than treated as context-free.
An output may be valid under context $c$ without being valid across all contexts.
Probabilistic weighting can represent uncertainty over alternatives; it is not itself semantic collapse. NRR instead specifies context-indexed records and explicit retention and commitment operations so that the status of alternatives is inspectable.

\subsection{Three Core Mechanisms}

NRR operates through three principles:

\textbf{(1) Context-indexed Non-Identity ($A \neq A$)}: Distinct occurrences of the same symbol may receive distinct representations across contexts.

\textbf{(2) Approximate Identity ($A \approx A$)}: Entities share gradient-based similarity, not binary sameness.
Similarity is continuous, allowing partial overlap without full identity.

\textbf{(3) Non-Resolution}: Conflicting interpretations can coexist without forced convergence.
Ambiguity is a valid computational state, not an error.

\section{Output-Level Pressures Toward Early Commitment}

\subsection{Behavior of the Tested Single-Embedding Baseline}

In the synthetic baseline evaluated here, one embedding is assigned per token type and the classifier is trained for a single label. Under the tested training setup, this configuration produces low Turn-1 output entropy.

\subsection{Energy Efficiency Arguments}

Retaining multiple interpretations appears computationally expensive.
If a system commits early and later context contradicts that commitment, revision or recomputation may be required. Whether this produces a net cost is an empirical question, not a result of the present study.
We hypothesize that NRR can become energy-positive in settings such as long multi-turn dialogues with frequent context shifts and creative generation, where the upfront cost of maintaining parallel interpretations may be offset by reduced revision or recomputation.

\section{NRR in Practice: Case Studies}

We illustrate how an NRR-style system could represent state in several scenarios. These are conceptual illustrations, not evaluations. Only the synthetic polysemy case in Section~\ref{sec:minimal_experiment} is tested empirically, and that experiment evaluates a gated Multi-Vector Embeddings-style component rather than the full architecture.

\subsection{Paradox Handling: Internal Representation Matters}

\textbf{Problem:} ``This sentence is false.''
 A system that fixes one truth value too early may fail to expose the competing readings relevant to the paradox. An NRR implementation would be designed to maintain contextual records (truth\_context, falsity\_context, paradox\_context) until a declared operation commits or reformulates them.
The paradox becomes \textbf{data} rather than \textbf{error}.

\subsection{Creative Generation: Polysemy in Writing}

\textbf{Task:} ``Write a story about `light'.''
A system that selects one sense early may produce a single-layer output. An NRR implementation would retain candidate meanings (illumination, weight, mood) in its externalized state so that later generation policies can inspect or combine them.

\subsection{Context-Dependent Identity}

\textbf{Problem:} Context shift in dialogue (e.g., ``The bank is solid'' $\to$ ``ducks are swimming'').
A system that fixes the financial reading may need to revise it after the second cue. An NRR implementation would retain financial and river records and reweight them when later context arrives.

\section{Implementation Framework}
\label{sec:implementation}

This section presents architectural components for implementing NRR.
While related to prior work on multi-prototype embeddings \cite{reisinger2010multi, levine2020sensebert, loureiro2022lmms} and non-softmax attention \cite{katharopoulos2020transformers, choromanski2021rethinking}, the mechanisms in NRR are unified toward explicit ambiguity preservation rather than sense selection or sparsity.

\subsection{Multi-Vector Embeddings}

Unlike the controlled and static token-type conventions that assign one vector per token type, NRR embeddings are designed to retain multiple contextual variants simultaneously.
\begin{equation}
\mathcal{H}_t = \{\mathbf{h}_{t,1}, \mathbf{h}_{t,2}, \ldots, \mathbf{h}_{t,k}\}
\end{equation}
These could be trained via contrastive initialization or adversarial diversity loss to ensure variants capture distinct semantic dimensions; we leave concrete training schemes to future empirical work.

\subsection{Non-Collapsing Attention}

Softmax normalizes attention weights competitively; that property alone does not establish semantic erasure. NRR explores independently activated weights (e.g., sigmoid or other non-competitive functions) as one explicit retention mechanism that allows multiple positions to receive strong activation simultaneously.

\begin{figure}[ht]
\centering
\begin{verbatim}
Algorithm: Non-Collapsing Attention Process

Input: Interpretation sets {H_t} for t=1 to n
Output: Updated interpretation sets {H'_t}

For each token t:
  For each interpretation i in {1...k}:
    1. Compute queries q, keys k, values v
    2. Compute scores: s_ij = q * k^T / sqrt(d)
    3. Apply Activation (No Softmax):
       alpha_ij = sigmoid(s_ij)
    4. Aggregate:
       h'_t,i = sum_j (alpha_ij * v_t,j)
  End
End
\end{verbatim}
\caption{Conceptual logic for Non-Collapsing Attention.
Unlike standard attention, scores are activated independently (e.g., via sigmoid) rather than competing via softmax.}
\label{fig:attention_logic}
\end{figure}

Stability could be maintained through techniques such as layer normalization \cite{ba2016layer} and residual scaling \cite{bachlechner2020rezero}, along with gradient clipping and attention strength regularization.

\subsection{Contextual Identity Tracking (CIT)}

CIT explicitly tracks which interpretation belongs to which semantic context.

\subsubsection{Context Detection}
When semantic discontinuity exceeds a threshold:
\begin{equation}
d(\mathbf{h}_t, \mathbf{h}_{t-1}) > \tau_{\text{context}}
\end{equation}
a new context ID is assigned.
$\tau_{\text{context}}$ may be learned or fixed; in a minimal implementation, it could be defined as the cosine-distance threshold separating intra-turn variability from inter-turn shifts.

\subsection{Strategic Resolution: When and How to Collapse}

While NRR treats non-resolution as computationally valid, practical systems must eventually act.
NRR separates three control terms:
\begin{itemize}
\item \emph{Collapse} is the turn-level selection or projection event required to produce a task-facing output.
\item \emph{Commit inertia} is the persistence of that selection across later updates when new evidence arrives.
\item \emph{Unmarked collapse} is the condition in which output selection is not distinguished from destruction of the retained alternatives.
\end{itemize}
NRR does not eliminate collapse events. It makes commitment timing explicit and keeps a
task-facing projection separate from the structured state that remains available for
revision. Resolution policies may depend on task affordances, context stability, and
interpretation dominance, but they remain \textbf{external to the NRR core}. A committed
output is produced when the current context supports resolution and action is required;
otherwise the state remains unresolved, including explicit conflict structure where
applicable. Later evidence may reopen a previously selected state.

\subsection{Protocol Extensions: State Space and Operators}
\label{sec:protocol}

The architectural components above (MVE, NCA, CIT) can be unified under a formal state space that enables systematic extension. This section provides the minimal specification for researchers implementing NRR-based systems.

\subsubsection{State Space Formalization}

The non-collapsing state is represented as:

\begin{equation}
\mathcal{S} = \bigl((v_i, c_i, w_i, m_i)\bigr)_{i=1}^{n}
\end{equation}

where $v_i \in \mathbb{R}^d$ is a semantic vector, $c_i$ is a context identifier, $w_i \in \mathbb{R}_{\geq 0}$ is a nonnegative activation weight, and $m_i$ is optional metadata. The Core--Phi contract does not require an upper bound on $w_i$; an implementation may use bounded scores as a special case. Crucially, the weights $w_i$ are \textit{not} normalized to sum to unity---multiple interpretations can maintain high activation simultaneously without competing for probability mass. This supports independently active records; retention additionally requires that multiple warranted records remain present and that subsequent operators do not delete them by default.

For measurement purposes, on states with positive total activation
$\sum_j w_j>0$, we associate a discrete distribution over interpretations:
\begin{equation}
p(i \mid \mathcal{S}) = \frac{w_i}{\sum_j w_j}
\end{equation}
and define $H(\mathcal{S})$ as the Shannon entropy of this distribution:
\begin{equation}
H(\mathcal{S}) = - \sum_{i=1}^{n} p(i \mid \mathcal{S}) \log_2 p(i \mid \mathcal{S})
\end{equation}
This normalization is used only for evaluation; the internal NRR dynamics do not require $w_i$ to sum to unity. Entropy is undefined for the all-zero state, which is therefore outside the entropy-evaluation domain.

\subsubsection{Interference Operators}

We introduce a candidate family of interference operators that act on $\mathcal{S}$.
The set is extensible rather than fixed; practical implementations may use a minimal subset.
Downstream implementations may instantiate this extensibility at state-update or policy level under separately declared protocols.
We \emph{define} an NRR interference operator as a transformation $T$ on $\mathcal{S}$
whose intended property is approximate information preservation, i.e., our design goal is that
\begin{equation}
H(T(\mathcal{S})) \geq H(\mathcal{S}) - \epsilon
\end{equation}
for small $\epsilon$.
This entropy criterion is evaluated only on positive-total input states for which
$T(\mathcal{S})$ also has positive total activation.
Table~\ref{tab:operators} specifies the operators and their roles at the candidate/specification level. A companion paper makes the record carry-forward and entropy-conformance layers precise and executable \cite{saito2026phi}.
\textbf{Notation note:} Later implementation papers distinguish state-level and item-level operators.
Those papers may use item-level updates (often denoted $\sigma_{\text{item}}, \delta_{\text{item}}$), while this table lists the broader state-level candidate family.
One operator, CPP Integration ($\kappa$), implements the Contradiction-Preservation Principle: contradictory hypotheses are retained rather than eliminated.

\begin{table}[h]
\centering
\small
\begin{tabular}{lll}
\toprule
\textbf{Operator} & \textbf{Symbol} & \textbf{Function} \\
\midrule
State-level Calibration & $\sigma$ & Uniform scaling/check operator (may be entropy-invariant) \\
Label-Free Abstraction & $\alpha$ & Projects to geometric relations without labels \\
Positioning & $\rho$ & Assigns contextual coordinates (implements CIT) \\
Invariance & $\iota$ & Identifies structure stable across contexts \\
Dampening & $\delta$ & Reduces activation differentials \\
Deferred Resolution & $\tau$ & Postpones projection to output boundary \\
CPP Integration & $\kappa$ & Preserves contradictory states with conflict tags \\
Persistence & $\pi$ & Maintains state across turns with decay \\
\bottomrule
\end{tabular}
\caption{NRR interference operators. Each is intended to transform $\mathcal{S}$ in ways that avoid collapse.}
\label{tab:operators}
\end{table}

One possible pipeline: $\mathcal{S}' = \pi(\kappa(\delta(\rho(\sigma(\mathcal{S}))), \mathcal{S}_{\text{new}}))$---stripping biases, positioning, dampening, integrating new information, and persisting.

\textbf{Note on operator constraints:} While NRR permits non-resolution as a valid state, not all transformations preserve this property. Operators must satisfy design principles (e.g., relative structure preservation) to avoid collapsing the very ambiguity NRR maintains. The companion contract paper provides one bounded formalization and conformance surface \cite{saito2026phi}; broader operator classes remain open.

\subsubsection{Non-Destructive Projection}

NRR produces singular output through a task-facing projection that leaves the specified retained state unchanged:

\begin{equation}
\Pi(\mathcal{S}) = g(v_{i^*}), \quad i^* = \arg\max_i w_i, \quad \mathcal{S}' = \mathcal{S}
\end{equation}

where $g: \mathbb{R}^d \to \mathcal{Y}$ maps the semantic vector to the output space (e.g., token vocabulary). The specification requires the retained state $\mathcal{S}$ to remain unchanged by projection. Whether existing systems preserve comparable alternatives across output is empirical and is not tested here.

\section{Controlled Retention-to-Resolution Experiment}
\label{sec:minimal_experiment}

\textbf{Scope of this section:} This controlled synthetic experiment measures output entropy for a gated Multi-Vector-Embeddings-style component and a controlled single-embedding baseline before and after a contextual cue. Its evidence applies to those tested configurations, not to Non-Collapsing Attention, CIT, the operator family, the complete NRR architecture, or modern LLMs.

\vspace{0.5em}

To test one operational consequence of the context-indexing convention, we ask: \textbf{does the tested output distribution retain uncertainty when context is insufficient?}

\subsection{Task Design: Turn 1 Entropy Measurement}

We created a synthetic two-turn disambiguation task where the identity of an ambiguous token (``bank'') is determined only after context arrives.
\begin{itemize}
    \item \textbf{Turn 1 (Ambiguous):} ``The bank is \{adj\}.'' (e.g., solid, stable, old, new)
    \item \textbf{Turn 2 (Context):} Disambiguating cue (e.g., ``investor'' $\to$ FINANCIAL, ``ducks'' $\to$ RIVER).
\end{itemize}

\textbf{Key Insight:} Rather than measuring classification accuracy (which depends on label distribution), we directly measure \textbf{entropy of the output distribution at Turn 1}---before any disambiguating context arrives.

\begin{equation}
H(p) = -\sum_{i} p_i \log_2 p_i
\end{equation}

For a binary classification:
\begin{itemize}
    \item $H_{\max} = \log_2 2 = 1.0$ bit (maximum binary entropy: $p = [0.5, 0.5]$)
    \item $H_{\min} = 0$ (complete certainty: $p = [1, 0]$ or $[0, 1]$)
\end{itemize}

A model that preserves ambiguity should maintain high entropy at Turn 1; a model that collapses early will show low entropy.

\subsection{Model Comparison}

\begin{itemize}
    \item \textbf{Baseline:} Standard single embedding per token, 32-dimensional embeddings, 2-layer MLP classifier.
    \item \textbf{NRR-lite:} Implements the MVE principle in minimal form, maintaining $k=2$ separate embeddings for the ambiguous token (``bank''). A context-dependent gate selects the appropriate embedding based on Turn 2 context. When Turn 2 is neutral (no context), the gate remains balanced.
\end{itemize}

Both models were trained on 1,000 samples with balanced labels (50\% FINANCIAL, 50\% RIVER) for 100 epochs.
All reported entropy values and statistics are computed from the trained models over 5 independent random seeds.

\subsection{Results}

\begin{table}[h]
\centering
\caption{Turn 1 output-distribution entropy in one controlled gated MVE-style
NRR-lite component and its single-embedding baseline (5 random seeds). Both
configurations achieve 100\% accuracy after disambiguating Turn 2 context.}
\label{tab:nrr_entropy}
\begin{tabular}{lccc}
\toprule
\textbf{Model} & \textbf{Turn 1 Entropy $H$ (bits)} & \textbf{Gate Entropy} & \textbf{Context Accuracy} \\
\midrule
Baseline       & $0.15 \pm 0.13$ & --- & 100\% \\
NRR-lite       & $\mathbf{0.91 \pm 0.04}$ & $1.00$ & 100\% \\
\midrule
Maximum $H$    & $1.00$ & $1.00$ & --- \\
\bottomrule
\end{tabular}
\end{table}

\begin{figure}[ht]
\centering
\includegraphics[width=\textwidth]{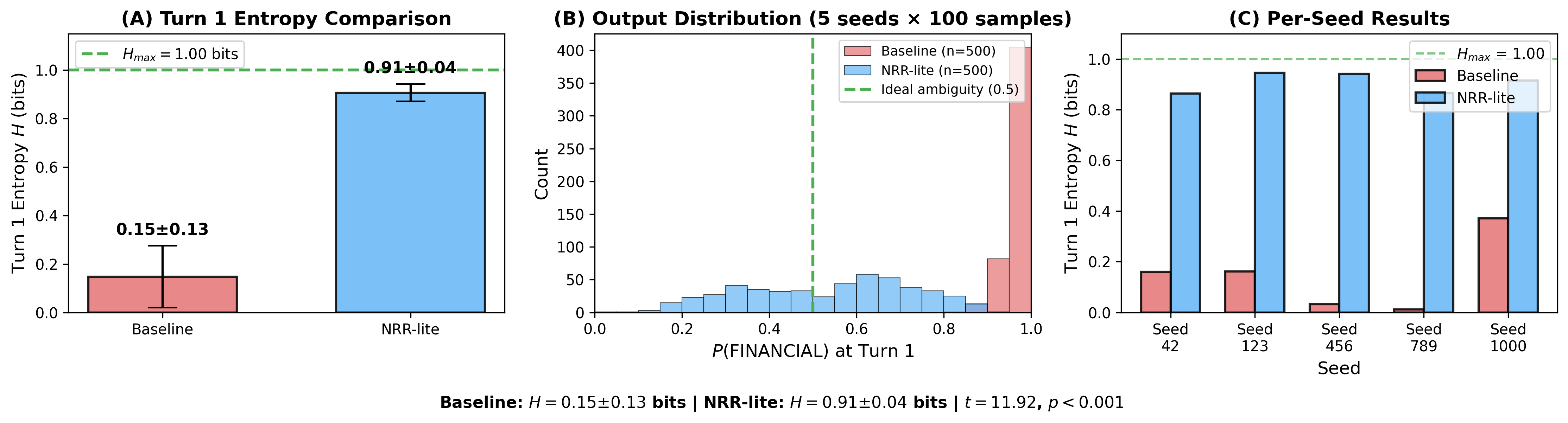}
\caption{Turn 1 output-distribution entropy for one controlled gated MVE-style
NRR-lite component and its single-embedding baseline (aggregated across 5 seeds);
both configurations achieve 100\% accuracy after disambiguating Turn 2 context.
\textbf{Left:} Entropy at Turn 1 (no context). NRR-lite produces high entropy ($H \approx 0.91$ bits), near the theoretical maximum, while the controlled baseline produces low entropy ($H \approx 0.15$ bits).
\textbf{Center:} Distribution of $P(\text{FINANCIAL})$ at Turn 1 across all test samples (500 total: 5 seeds $\times$ 100 samples each). NRR-lite centers around 0.5 (ambiguous); Baseline clusters near 1.0 (committed to financial interpretation).
\textbf{Right:} Per-seed Turn 1 Entropy breakdown across all 5 random seeds: NRR-lite (blue) remains near $H \approx 0.9$ bits, while the Baseline (red) ranges from 0.01 to 0.37 bits.}
\label{fig:nrr_experiment}
\end{figure}

\textbf{Key Findings:}
\begin{enumerate}
    \item \textbf{Baseline Output ($H = 0.15$ bits):} Under the tested setup, the controlled single-embedding baseline assigns a low-entropy output distribution at Turn 1 despite the absence of disambiguating context.
    \item \textbf{NRR-lite Output ($H = 0.91$ bits):} The gated multi-vector configuration maintains near-maximum output entropy at Turn 1 under the same setup.
    \item \textbf{Gate Behavior:} NRR-lite's context gate achieves $H = 1.0$ bit (perfect balance) when no context is provided, shifting appropriately when Turn 2 arrives.
    \item \textbf{Resolution Capability in This Setup:} Both models achieve 100\% accuracy with context, so high Turn-1 entropy and accurate contextual resolution co-occur in the tested configuration.
    \item \textbf{Across-Seed Consistency:} The entropy difference appears across all 5 random seeds (paired $t$-test: $t(4) = 11.9$, $p < 0.001$).
    \item \textbf{Seed Sensitivity:} Across the five tested seeds, baseline entropy ranges from 0.01 to 0.37 bits ($\text{SD} = 0.13$), while NRR-lite entropy varies less ($\text{SD} = 0.04$), quantifying variability within this setup.
\end{enumerate}

The tested gated multi-vector configuration therefore exhibits the complete measured behavior targeted by this experiment: high Turn-1 output entropy before context and accurate resolution after context. This reproducible result supplies a concrete implementation point for the broader retention--commitment interface specified in the paper. Its scope is the tested component and configuration described above.

\section{Future Evaluation Framework}

Comprehensive evaluation requires scaling beyond this minimal verification.
We propose benchmarks for:
\begin{itemize}
    \item \textbf{Ambiguity Retention:} Measuring entropy of interpretations under polysemy.
    \item \textbf{Context-Shift Adaptation:} Measuring cost of adaptation vs. backtracking.
    \item \textbf{Creative Generation:} Human evaluation of semantic richness.
    \item \textbf{Paradox Stability:} Testing system stability under self-reference.
\end{itemize}

\paragraph{Reproducibility note.}
We release the implementation, fixed experiment settings, and primary execution logs in the companion repository.
Core controls are fixed at run time, including model configuration, random seeds, and trial count (temperature is not used in this implementation).
The main paper artifact is generated by the multi-seed command and written as a machine-readable results file.
Exact environment details and command-to-artifact mapping are documented in \texttt{nrr-core/reproducibility.md}.

\section{Conclusion}

This paper specified \textbf{Non-Resolution Reasoning (NRR)}, a state-level framework that makes context-indexed retention and commitment explicit and inspectable.
In one controlled synthetic setting, a gated multi-vector configuration maintains high output entropy ($H = 0.91$ bits) when context is absent, while the controlled single-embedding baseline has low output entropy ($H = 0.15$ bits); both achieve perfect accuracy when context arrives. The result demonstrates, in the tested configuration, that high pre-context output entropy can coexist with accurate later resolution.

The paper's central contribution is the retention--commitment interface together with this reproducible implementation point. The experiment does not establish full-architecture validation, a universal causal mechanism, or matched-parameter superiority; those questions require broader comparisons than the present component study.

NRR is therefore not anti-commitment. It makes commitment a declared operation whose timing can be designed, inspected, and revised while the retained alternatives remain computationally available.
\textbf{The question is not whether AI should resolve ambiguity, but when, how, and under whose control.}

\section*{Acknowledgments}
The author acknowledges the use of large language models, including Claude (Anthropic), ChatGPT and Codex (OpenAI), and Gemini (Google), for language editing, proofreading, and LaTeX formatting assistance during manuscript preparation. All substantive ideas, claims, analyses, and conclusions are solely the responsibility of the author.

\end{document}